\documentclass[a4paper,twoside]{article}

\usepackage{epsfig}
\usepackage{subcaption}
\usepackage{calc}
\usepackage{amssymb}
\usepackage{amstext}
\usepackage{amsmath}
\usepackage{amsthm}
\usepackage{multicol}
\usepackage{pslatex}
\usepackage{apalike}
\usepackage[bottom]{footmisc}
\usepackage[sectionbib, round]{natbib}
\PassOptionsToPackage{hyphens}{url}\usepackage{hyperref}
\usepackage{SCITEPRESS}

\begin{document}

\title{Two-Step Data Augmentation for Masked Face Detection and Recognition: Turning Fake Masks to Real}

\author{\authorname{Yan Yang Aaren\sup{1}, George Bebis\sup{1} and Mircea Nicolescu\sup{1}}
\affiliation{\sup{1}Department of Computer Science and Engineering, University of Nevada, Reno, 1664 N Virginia St, Reno, NV 89557, USA}
\email{yanyang@nevada.unr.edu, \{bebis, mircea\}@cse.unr.edu}
}


\abstract{The COVID-19 spread raised urgent requirements for masked face recognition and detection tasks. However, the current masked face datasets are insufficient. To alleviate the limitation of data, we proposed a two-step data augmentation that combines rule-based mask warping with unpaired image-to-image translation. Our qualitative evaluations showed that our method achieved noticeable improvements compared to the rule-based warping alone and complemented results from other state-of-the-art GAN-based generation methods, such as IAMGAN. The non-mask change loss and the noise input we used to improve training showed effectiveness. We also provided an analysis of potential future directions based on observations of our experiments.}

\onecolumn \maketitle \normalsize \setcounter{footnote}{0} \vfill

\section{\uppercase{Introduction}}
\label{sec:introduction}

Computer Vision tasks like recognition, detection, classification, etc., performed on occluded human faces, existed even before the COVID-19 outbreak \citep{8099536}. The spread of COVID-19 has urgently imposed performance and robustness requirements to such applications. However, multiple latest masked-face detection or recognition works \citep{singh2021face, DBLP:journals/corr/abs-2104-09874} claim their models only as starting points of future transfer learning with more data, instead of final results, or at least recognize the data insufficiency problem. The research community has built mature datasets of dominantly full faces, but masked face datasets are still under construction.

For full faces, recognition tasks use single-face image sets, with every single identity assigned to multiple facial images \citep{8614364}, and detection algorithms work on multi-face scene images and learn bounding box locations together with an optional class label for each box \citep{fddbTech, 8569109}. To improve the masked-face datasets to facilitate the same learning approaches, we either collect and annotate raw data or generate artificial images to augment existing data \citep{2020}. In this paper, we focus on generating artificially masked faces.

Recent works address this problem by modifying unmasked faces into masked. Some use rules to warp masks onto faces \citep{wang2020masked, DBLP:journals/corr/abs-2008-11104, cabani2021maskedface}. One uses Neural Network (NN) to translate unmasked faces into masked \citep{geng2020masked}. The rule-based methods provide realistic mask textures and completely avoid the risk of distorting other parts of faces. However, they often result in bad transitions between masks and background faces, and the lighting on masks often looks unreal. Some rule-based algorithms \citep{DBLP:journals/corr/abs-2008-11104, DBLP:journals/corr/abs-2109-05804} achieved mask diversity by defining different mask image options, but this diversity is limited to the predefined mask types. In contrast, NN methods learn to avoid facial distortions and generate mask textures, often to a reasonable extent but never with promise. They provide natural transition, realistic details, and sometimes more diversity in compensation.

We propose a two-stage approach combining a rule-based method and image-to-image translation (I2I). After applying rule-based methods to full faces, we use an I2I model to translate rule-generated masks into more realistic ones. Rule-generated mask regions are calculated to serve as ground truth attention areas, from which we designed an extra loss to restrict I2I modifications only to mask regions. The rest of the paper will call the raw data ``full-face'' images, the faces with rule-based masks ``rule-based mask'' images, and the final outputs ``realistic mask'' images.

The applications of fake mask methods include not only masked face recognition/classification but also detection that requires multi-face images. By extracting bounding boxes in multi-face images, converting a portion into masked, and overlaying masked faces on original boxes, we transform multi-face images to serve masked face detection training.

\section{\uppercase{Related Work}}

\noindent {\bf Real-World Datasets.}
Traditional full-face datasets embody a large number of faces with high variations in demographics, head rotations, facial landmarks, occlusion degrees, facial expressions, etc. \citep{SAGONAS20163, DBLP:journals/corr/abs-1812-04948} \footnote{Occlusions here include various types like body parts, scarves, etc. Medical masks constitute only a tiny portion.} Annotations to these datasets include categories mentioned above, as well as subject identities for facial recognition, bounding boxes for facial detection, and so on.

On the other hand, datasets emphasizing masked faces are limited in the quantity of data and the variation of features. \citet{wang2020masked} developed two different datasets based on real-world images. Real-World Masked Face Dataset (RMFD) \citep{wang2020masked} for recognition tasks contains 5,000 masked and 90,000 normal faces belonging to 525 people. Masked Face Detection Dataset (MFDD) \citep{wang2020masked} contains 24,711 masked face images for detection tasks, which is currently not publicly accessible \footnote{\url{https://github.com/X-zhangyang/Real-World-Masked-Face-Dataset/issues/16}}. MAsked FAces (MAFA) by \citet{8099536} with 30,811 internet images and 35,806 masked faces is the largest real-world masked face dataset to our best knowledge. It is annotated with relatively abundant information such as face bounding boxes, mask bounding boxes, mask types, face orientations, occlusion degrees, gender, race, and more, but MAFA alone is not comparable with all the diversity of currently available full-face datasets. Moreover, no identity information is provided to MAFA, so it is only for face detection tasks. Besides RMFD, recognition tasks may also use Masked Faces in Real World for Face Recognition (MFR2) \citep{DBLP:journals/corr/abs-2008-11104}, a small set of 269 images belonging to 53 identities. \citet{geng2020masked} proposed a two-part Masked
Face Segmentation and Recognition (MFSR) dataset. 9,742 masked images were collected from the web and annotated with mask regions in the first part. The second part contains 11,615 faces, masked or non-masked, for 704 real-world identities and 300 internet-obtained identities.

\noindent {\bf Artificially Masked Faces.} Some researchers generated artificial data to augment existing real-world masked face datasets. Stemmed from RMFD and MFDD, the same lab proposed Simulated Masked Face Recognition Dataset (SMFRD) \citep{wang2020masked}, using a naive copy-and-paste method to put cartoon mask images onto existing face recognition datasets. \citet{DBLP:journals/corr/abs-2008-11104} and \citet{cabani2021maskedface} separately used more sophisticated methods to warp mask images onto faces based on detected facial landmarks \citep{SAGONAS20163}. \citet{DBLP:journals/corr/abs-2008-11104} provided their method as a MaskTheFace tool for both single- and multi-face images. \citet{cabani2021maskedface} published a single-face dataset, MaskedFace-Net, applying their rule-based method to Flickr-Faces-HQ \citep{DBLP:journals/corr/abs-1812-04948}. MaskedFace-Net consists of Correctly and Incorrectly Masked Face Dataset (CMFD and IMFD) with about 70,000 synthesized images each. CMFD is used as our model input. While our paper was in progress, \citet{DBLP:journals/corr/abs-2109-05804} published MLFW (Masked LFW), which enhanced the landmark-based warping by extra rule-based improvements on the unreal lighting and abrupt mask boundaries.

Beyond rule-based methods, \cite{geng2020masked} proposed an NN model, Identity Aware Mask GAN (IAMGAN), to synthesize masked faces. It consists of a CycleGAN-like generation module and an Identity Preservation (IP) module. The IP module has a mask region predictor and an identity classifier. The former predicts mask regions and removes them from both masked (output) and unmasked (input) images. The latter classifies identities using the rest parts and penalizes identity change between input and output based on information from multiple classifier levels.

\noindent {\bf GAN Models for I2I.} Image to image translation is a problem of mapping images from one domain into another, with the intrinsic source content preserved and the extrinsic target style transferred \citep{DBLP:journals/corr/abs-2101-08629}. It has been used in a broad set of scenarios such as image synthesis \citep{DBLP:journals/corr/abs-1803-03396}, image segmentation \citep{GUO2020127}, style transfer \citep{DBLP:journals/corr/ZhuPIE17}, and more.
GAN models are adapted to the I2I tasks by replacing the generator's standard random input with source image embeddings and keeping the discriminator's function of distinguishing synthesized and authentic images.

The earliest versions of GANs for image translation, such as pix2pix \citep{DBLP:journals/corr/IsolaZZE16}, require aligned image pairs as the training data. However, getting a large number of strictly aligned image pairs is hard or even impossible in many situations. \citet{DBLP:journals/corr/ZhuPIE17} proposed CycleGAN using a cyclic loss, making it possible to train an I2I on unpaired data. The previously mentioned IAMGAN \citep{geng2020masked} is an enhanced CycleGAN aiming specifically to generate masks. Based similarly on CycleGAN, \citet{tang2021attentiongan} trained the generator to produce attention masks together with the generated image contents, which guided the generator to translate individual objects without changing the background. They call their general-purpose model ``AttentionGAN''.

Our work adapts AttentionGAN \citep{tang2021attentiongan}. The key innovation is that our source images are not full-face but rule-based mask images, i.e., CMFD \citep{cabani2021maskedface}. We use a rule-based method as the first step and the adapted AttentionGAN as the second step. We show that: 1. warping mask images onto full faces provides style guidance and ground truth attention for better GAN model results; 2. the GAN model renders the rule-based results into more realistic details. With the latest advances in rule-based methods such as \citet{DBLP:journals/corr/abs-2109-05804}, our two-step proposal may still improve details.

\section{\uppercase{Data}}\label{sec:data}

Like other unpaired I2I models, both training and testing of AttentionGAN require two sets of data, A and B, A being the source and B being the destination.

\begin{figure}
    \centering
    {\epsfig{file = 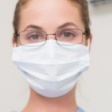, width = 1.4cm}}
    {\epsfig{file = 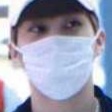, width = 1.4cm}}
    {\epsfig{file = 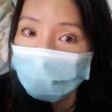, width = 1.4cm}}
    {\epsfig{file = 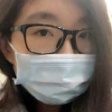, width = 1.4cm}}
    {\epsfig{file = 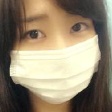, width = 1.4cm}}
    \caption{Examples from dataset B.}
    \label{fig:set_b}
\end{figure}

\noindent{\bf Dataset B.} We manually cropped web images and extracted MAFA bounding boxes for dataset B. From MAFA, we first extracted 8938 single faces with ``simple'' masks, full occlusions, sizes of at least 60 $\times$ 60 pixels, and front-facing orientations. Finding that annotations for MAFA are not accurate, we followed it by hand-picking a subset that strictly matches the criteria mentioned above. At the same time, we added ``faces without pitch/roll with light-colored medical masks'' as additional criteria. In this way, we have a controlled dataset for a more accessible proof of our concept that a superimposed mask image helps the subsequent I2I step, and they together achieve better results than the superimposing alone. With 1597 final images from MAFA, we supplemented the small subset with 98 cropped faces from open-source photos on \url{https://unsplash.com/} \footnote{Images credit to Jana Shnipelson et al. on \href{https://unsplash.com}{Unsplash}}, resulting in a total of 1695 images for training set B. Example images from B are shown in Figure \ref{fig:set_b}

\noindent{\bf Dataset A.} We use down-sampled CMFD as set A, which uses uniformly blue medical masks with rare occasions of misplacement. The downsampling makes sure the size of set A matches that of set B.

\section{\uppercase{Method}}

\begin{figure*}[!ht]
  \centering
   {\epsfig{file = 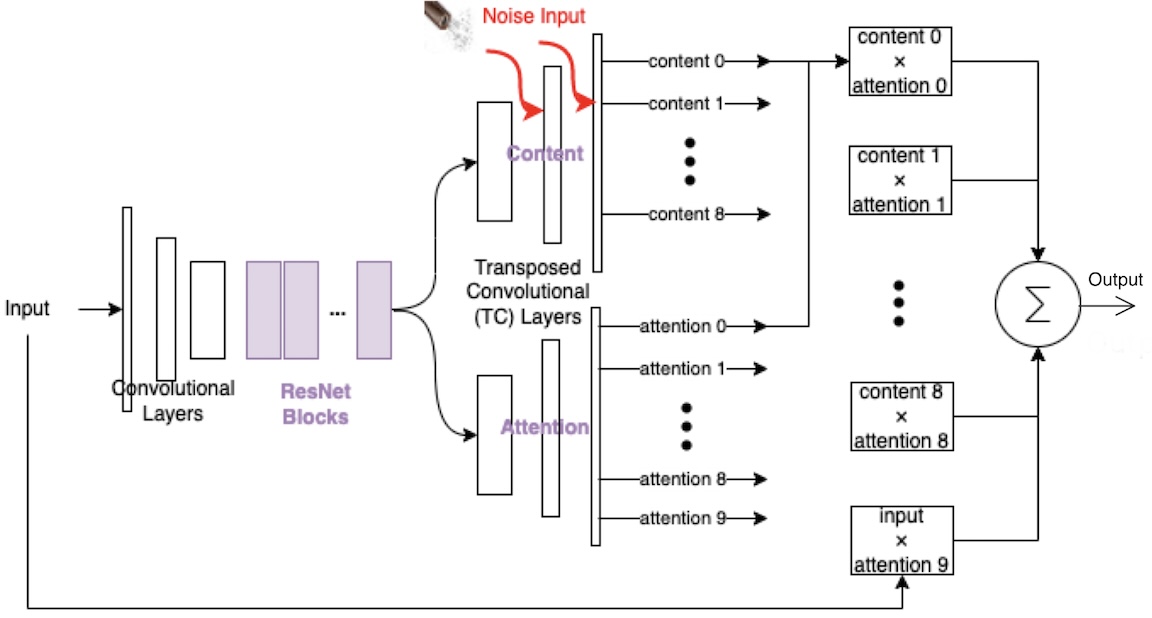, width = 11.5cm}}
  \caption{Generator architecture in AttentionGAN. Our noise inputs are depicted in red.}
  \label{fig:arch}
\end{figure*}

AttentionGAN is designed based on CycleGAN and shares with CycleGAN the co-training of translation models in two opposite directions. We use most of the default parameters in the AttentionGAN repository \footnote{\url{https://github.com/Ha0Tang/AttentionGAN}}, with batch size 4, learning rate 0.0002, Adam momentum 0.5, and weights initialized from Gaussian distribution $w\sim \mathcal{N}(0.0,\ 0.02)$.

Our input images are all resized to 256 $\times$ 256 with no cropping before being fed into the models. In each translation direction, our generator starts with a 3-pixel Reflection Padding followed by three convolutional layers with instance normalizations and nine ResNet blocks. In the convolutional layers, we increase the number of channels at each layer, and in the ResNet blocks, we keep the number of channels unchanged. Still in the generator, content tensors and attention masks are generated after the nine ResNet blocks in two separate pipelines, each of which consists of three Transposed Convolutional (TC) layers with two instance normalizations, where the number of channels increases layer-wise. The content tensors are activated by tanh and attention masks by softmax.

As shown in Figure \ref{fig:arch}, the generator produces corresponding attention masks for both output content tensors and the input image tensor. The tensor values are filtered by their attention masks and summed up to produce the final generator result.

On the other hand, the discriminator is a three-layer PatchGAN with kernel size 4 $\times$ 4, judging the input image's realness based on whether each 4 $\times$ 4 patch looks real. While retaining most of AttentionGAN's structures, we used multiple ways to adapt them to our training needs.

\subsection{Adding Non-Mask Change Loss}\label{sec:penalty}

As mentioned, the generator in AttentionGAN outputs a set of attention masks, which learn to find the most distinguishing parts between sets A and B unsupervised. However, we obtained sets A and B from different sources, causing heterogeneity beyond the facial mask differences. Examples include 1. faces in set B are more zoomed-in, in most extreme cases with foreheads partly cut out; and 2. Set B includes more outdoor scarves and hoods, which often occlude the lower parts of the masks and the foreheads.
These additional sources of heterogeneity caused the generator to produce inaccurate attention. On the other hand, the inputs for our GAN model are generated by warping mask images onto full faces, so the mask regions can be determined by comparing the full-face images with rule-based mask images pixel by pixel. Utilizing this pixel-by-pixel comparison to improve the misled attention became one of our improvement directions.

To achieve this, we created an extra ``Non-Mask Change (NMC)'' loss that calculates the L1 distances between the rule-based mask images and the realistic mask images for all pixels outside the mask regions. We minimize the sum of this L1 distance and other losses in AttentionGAN. The non-mask region for each rule-based mask image is calculated pre-training as a 256 $\times$ 256 boolean tensor, stored with the same file name as the rule-based mask image but with a different extension name. During training, these tensor files are paired with their rule-based mask images and the generated realistic mask images for the NMC loss calculation. Note that we only calculate the loss at the training stage, so we do not have to get these tensor files for any generalized model test or usage once we finish training. Our model learns to automatically generate the attention mask instead of relying on ground truth attention masks beyond training.

\subsection{Adding Noise}\label{sec:noise}

\begin{figure}[!ht]
  \centering
   {\epsfig{file = 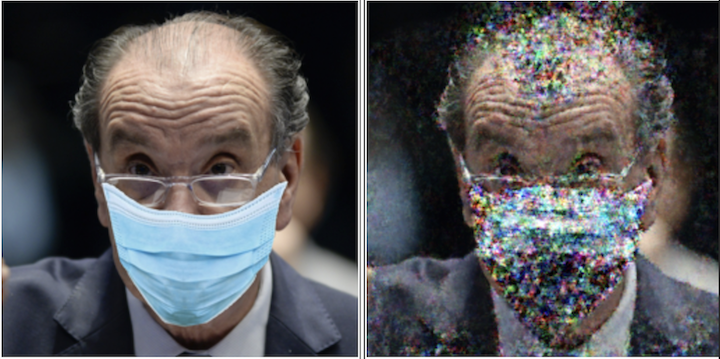, width = 2.95cm}}
  \caption{Feed noise to the last two content-generating layers without training.}
  \label{fig:test_noise}
\end{figure}

Inspired by StyleGAN \citep{DBLP:journals/corr/abs-1812-04948}, which takes random noise input to multiple generator layers for result diversity, we modified AttentionGAN to include zero-mean Gaussian noise input of a similar style.
We first attempted noise input to the first content-generating TC layer following the nine ResNet blocks. We tested the noise-tuned model structure on parameters trained without noise and found that the generated images remained the same no matter how we amplified the noise. We concluded that the first TC layer assimilates noise and does not map noise input to identifiable features in the final output.
Then, with one TC layer noise-tuned at a time, we tested different amplitudes of noise input on the other two content-generating TC layers, using the same model mentioned above. We found that noise added to either of the other two TC layers makes a difference to the model output, producing random noise pixels as shown in Figure \ref{fig:test_noise}.

Based on these tests, we trained our final model using a modified structure: the generator takes zero-mean Gaussian noise inputs to the last two content-generating layers, with standard deviations being 1 and 0.2 separately for the second and the last content-generating layers. We denoted our noise input location in Figure \ref{fig:arch}.

\begin{figure}[!ht]
  \centering
   {\epsfig{file = 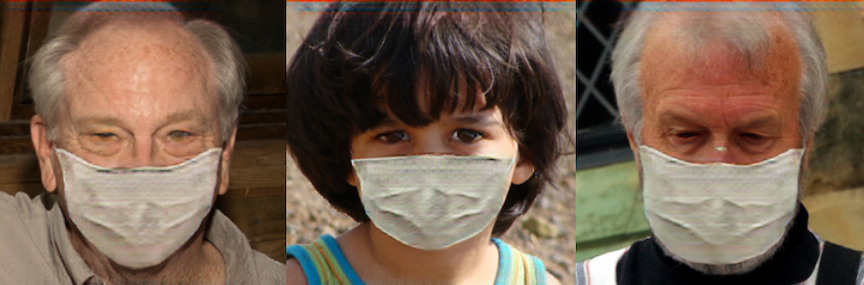, width = 4.9cm}}
  \caption{Uniform-colored masks generated at epoch 279.}
  \label{fig:uniform_color}
\end{figure}

Section \ref{sec:epochs} shows that models at two different epochs generated diverse colors after adding the noise input. In contrast, before adding the noise, our model output is in uniform color for any single checkpoint, as shown in Figure \ref{fig:uniform_color}. Therefore, we conclude that the noise input effectively results in increased diversity.

\begin{figure}
    \centering
    {\epsfig{file = 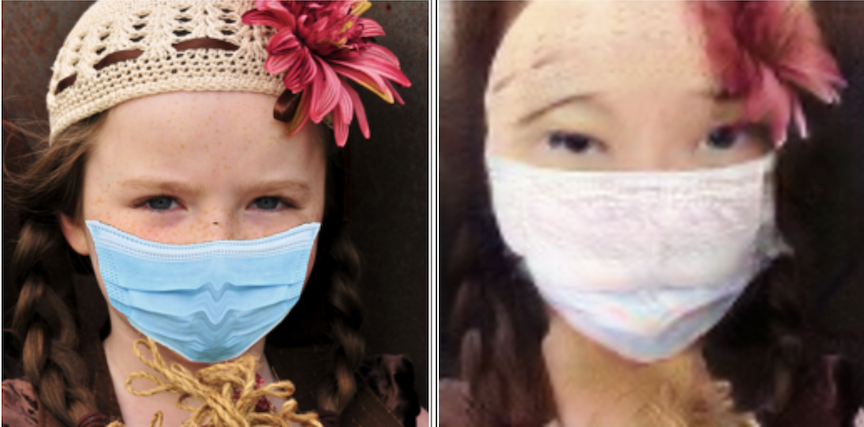, width = 2.9cm}}
    {\epsfig{file = 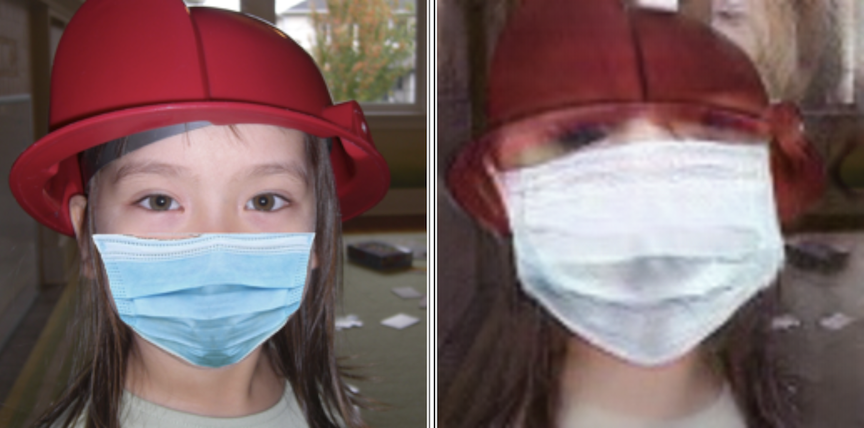, width = 2.9cm}}
    \caption{Models drastically distort inputs at two different epochs. Outputs like images 2 and 4 intermittently appear as the training progresses.}
    \label{fig:crash}
\end{figure}

Besides diversity, the noise input, together with the NMC loss, also reduced distortions to non-mask areas and stabilized training. Before the improvements, with the training process producing one sample per epoch, the samples did include faces mostly preserved from the input for consecutive epochs. However, drastically distorted and completely redrawn faces, as in Figure \ref{fig:crash}, also last for epochs often, interweaving with the preserved faces. In contrast, all epoch samples stably retained input faces after applying the improvements, with only local changes fluctuating, proving the increased training stability. The remaining non-mask changes still exist, with examples shown in Figure \ref{fig:476_noise}, but both the extent of changes and the stableness of training were improved.

\subsection{Transfer Learning and the Training Timeline}\label{subsec:transfer_learning}

We used transfer learning from our trial-and-error experiments. We gradually improved datasets and methods in our training timeline, but we did not discard previous checkpoints. Instead, we believe that a checkpoint from training epochs with less ideal model settings and datasets is better than a random start.

Our initial parameters for the single-face mask generation task are from a different research topic: a multi-face image translator trained with the same AttentionGAN model. These initial weights were gained by 60 epochs of training on Face Detection Data Set and Benchmark \citep{fddbTech} and MAFA data, with hyperparameters in the original AttentionGAN code set to $lambda_A = 5$, $lambda_B = 5$, and $lambda_{identity} = 0.2$. From each of FDDB and MAFA, about 25,000 images were selected as training sets A and B separately. We will not elaborate on the rationale behind our choices since it belongs to other research. The information here is only for the reproduction of our result.

Within the single-face mask generation task itself, our first experiment used all the 9,517 face images extracted from the MAFA training set, which were annotated as 1) fully occluded (with occlusion degree equal to or higher than three), 2) of ``simple'' mask type, and 3) at least 60 $\times$ 60 in size. From the analysis of this first experiment, we found our training set A, i.e., the CMFD data, is much less diverse in orientations of faces than set B. To prevent set B's additional orientation variations from confusing the model, we limited the image extraction to only facial orientations of ``front,'' ``front left,'' and ``front right,'' getting 8,938 images for set B. We downsampled set A to match the number. Our continued training used these 8,938-image datasets, with starting weights copied from the model at epoch 60 in our first experiment.

Then we noticed that faces in MAFA also involve pitch and roll rotations, which, unlike yaw rotations, are not annotated. On the other hand, CMFD mostly restricted pitch and roll rotations. Also, masks annotated as ``simple'' in set B are not always simple medical masks similar to CMFD. Cloth or gauze veils are also annotated as simple types. We believe limiting the source and destination datasets to have similar variations in the aspects above may help the model focus on the target modifications, i.e., the masks. Therefore, out of the 8,938 images, we manually selected 1,597, which are limited in pitch and roll rotations with only light-colored medical masks.

Adopting 98 additional real-world masked faces from online for set B, we finalized our datasets with 1,695 images in each set as described in Section \ref{sec:data}. After this, training and improvements, including in Sections \ref{sec:penalty} and \ref{sec:noise}, are all based on 1,695 real-world images and 1,695 CMFD images. The whole training timeline on the single face task is shown in Table \ref{tab:timeline}.

The training model was updated along the way as we designed new improvements, but training epochs on old models were utilized and stacked together. We might have tried more clear-cut experiments if we had time, but this methodology has accelerated training and alleviated that our final datasets are small.

\begin{table*}[!ht]
\caption{Transfer learning from trial-and-error experiments. All epochs use Learning Rate 0.0002 and Lambda\_identity 0.5.}\label{tab:timeline} \centering
\begin{tabular}{|c|c|c|c|c|l|}
 \hline
 Epochs
 & \multicolumn{1}{|p{2cm}|}{\centering Lambda$_A$ \\ /Lambda$_B$ }
 & \multicolumn{1}{|p{1cm}|}{\centering Noise \\ Input}
 & \multicolumn{1}{|p{2cm}|}{\centering Non-Mask \\ Change Loss}
 & \multicolumn{1}{|p{1cm}|}{\centering Dataset \\ Size}
 & \multicolumn{1}{|p{5cm}|}{\centering Training Data Selection \\ Restrictions Added to B}\\
 \hline\hline
 1$\sim$60 & 10/10 & No & No & 9,517 & Simple mask, fully occluded\\
 \hline
 61$\sim$90 & 10/10 & No & No & 8,938 & Front facing \\
 \hline
 91$\sim$140 & 8/8 & No & No & 1,695 & No pitch/roll, light-colored medical mask \\
 \hline
 141$\sim$298 & 8/8 & No & Yes & 1,695 & None\\
 \hline
 299$\sim$510 & 8/8 & Yes & Yes & 1,695 & None\\
 \hline
\end{tabular}
\end{table*}

\section{\uppercase{Discussions}}\label{sec:discussions}

Output from training epochs showed that our model slowly converged after applying noise input and NMC loss. We carefully watched this trend and visually picked two better-performing checkpoints, checkpoints 313 and 476, from the later epochs for our test.

\subsection{Improvements on Top of CMFD}\label{sec:epochs}

\begin{figure*}[!ht]
  
  \centering
   {\epsfig{file = 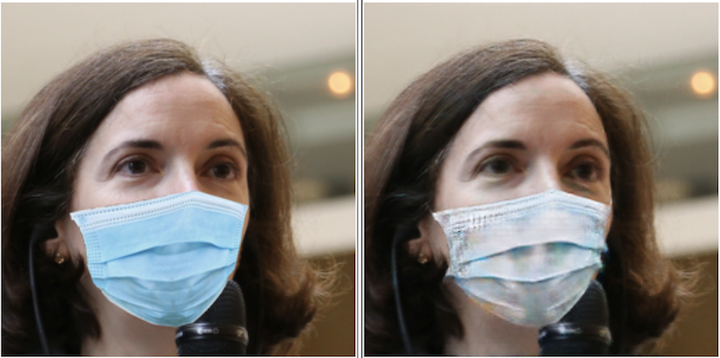, width = 4.4cm}}
   {\epsfig{file = 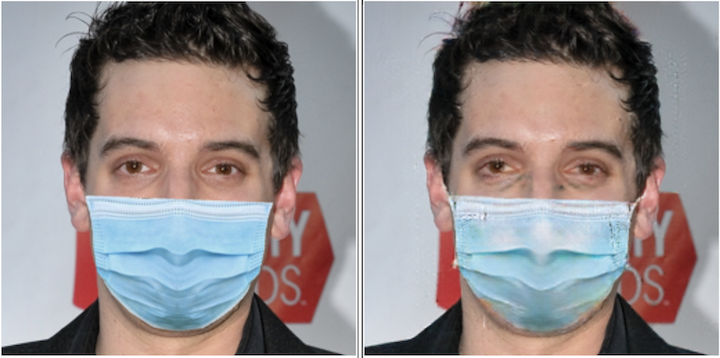, width = 4.4cm}}
   {\epsfig{file = 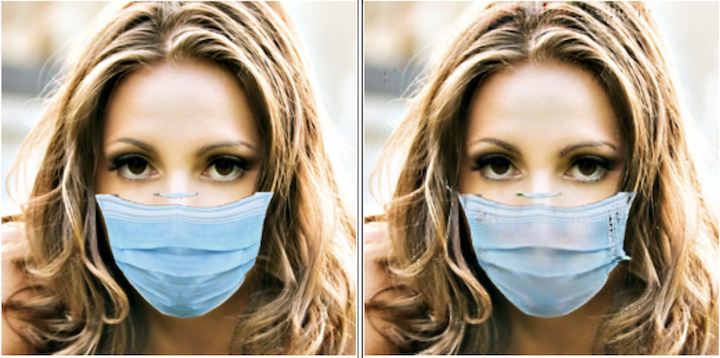, width = 4.4cm}}
   {\epsfig{file = 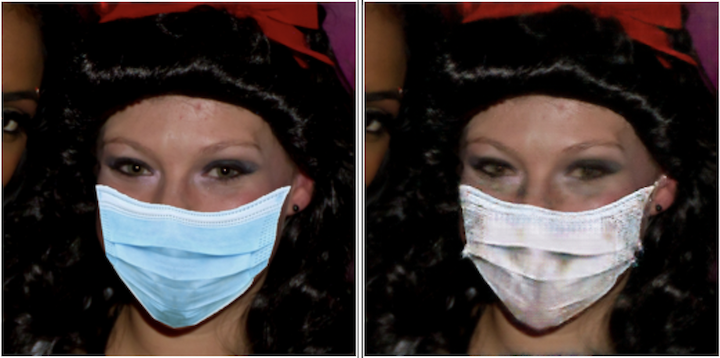, width = 4.4cm}}
   {\epsfig{file = 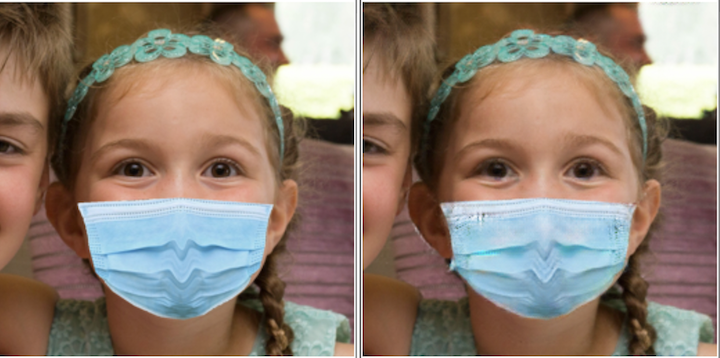, width = 4.4cm}}
   {\epsfig{file = 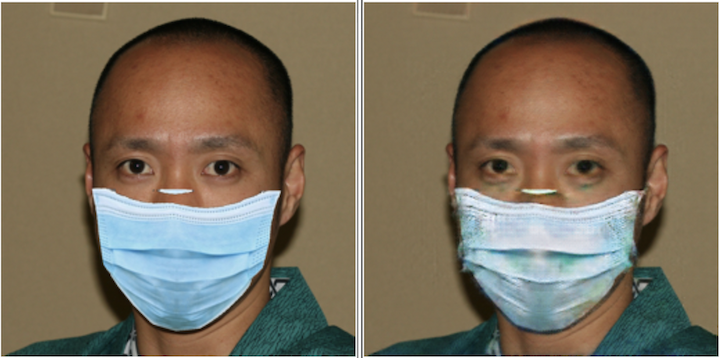, width = 4.4cm}}
   {\epsfig{file = 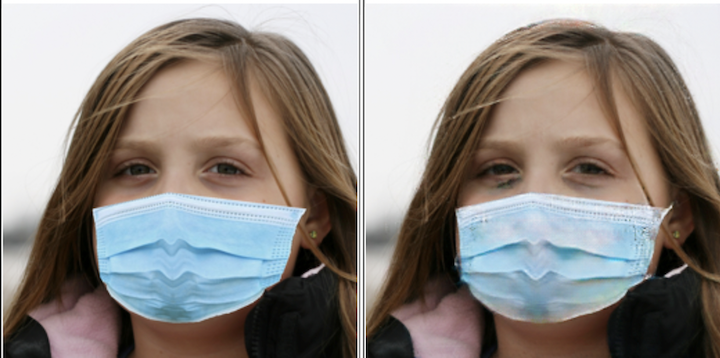, width = 4.4cm}}
   {\epsfig{file = 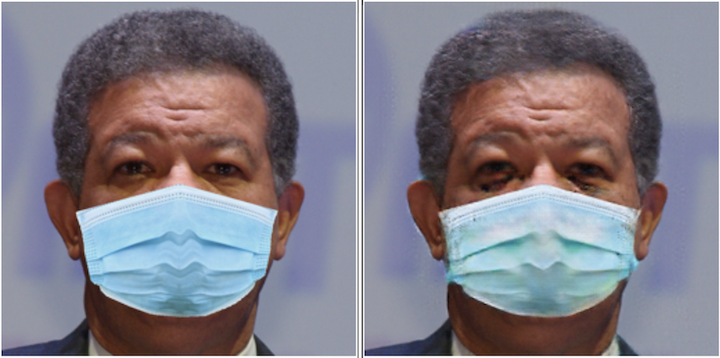, width = 4.4cm}}
   {\epsfig{file = 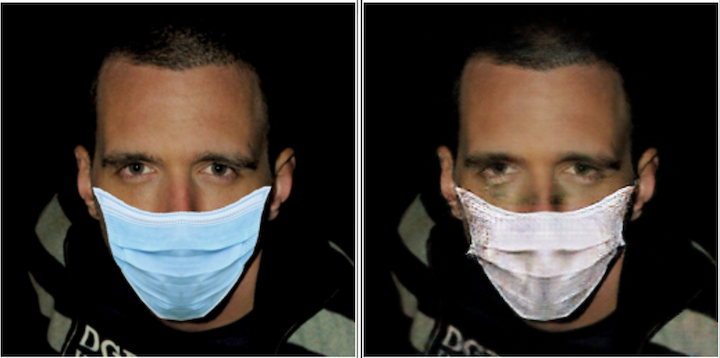, width = 4.4cm}}
  \caption{Results produced by the model at epoch 313. Input and output images are paired side by side.}
  \label{fig:epoch313}
\end{figure*}

Test results in Figure \ref{fig:epoch313} show that, compared to CMFD inputs, epoch 313 provides a diversity of mask colors that match dataset B's color distribution. It also shows better details than CMFD on various aspects:

\begin{itemize}
    \item Fabric folds and resulting irregular mask region boundaries;
    \item Straps or their connecting points with the masks;
    \item More realistic lighting matching cheek curvatures;
    \item Visual effects of masks lifted by the nose bridges;
    \item More natural transitions from masks to faces.
\end{itemize}

\begin{figure*}[!ht]
  
  \centering
   {\epsfig{file = 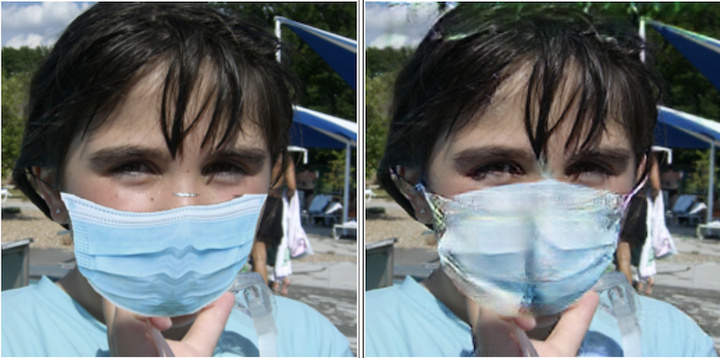, width = 4.4cm}}
   {\epsfig{file = 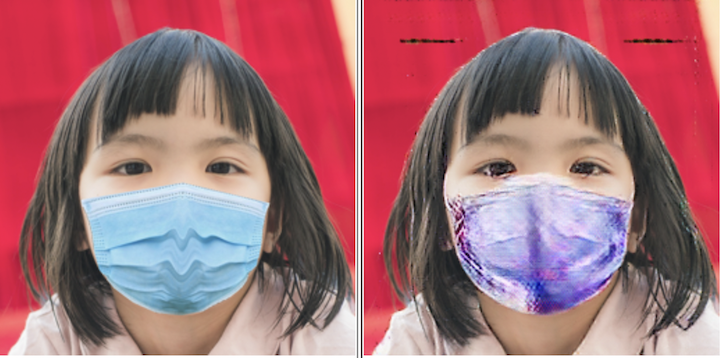, width = 4.4cm}}
   {\epsfig{file = 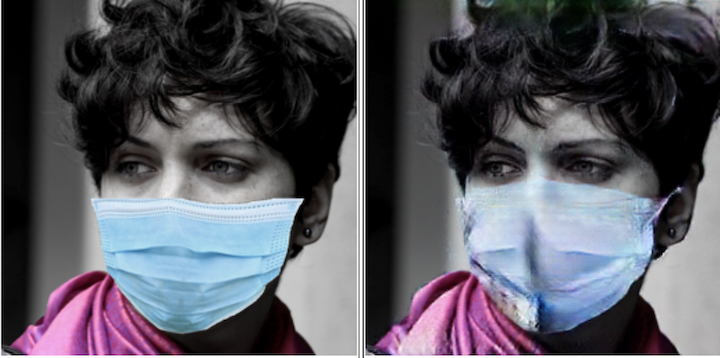, width = 4.4cm}}
   {\epsfig{file = 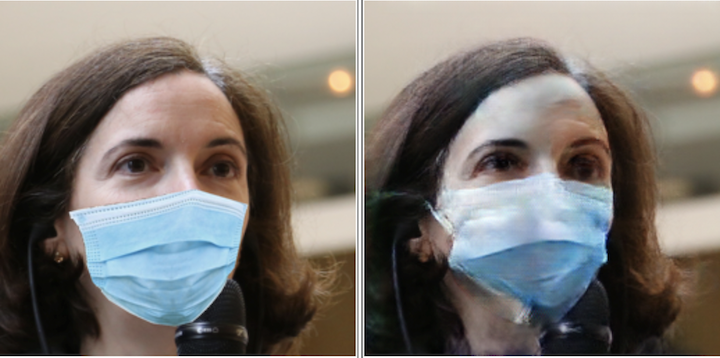, width = 4.4cm}}
   {\epsfig{file = 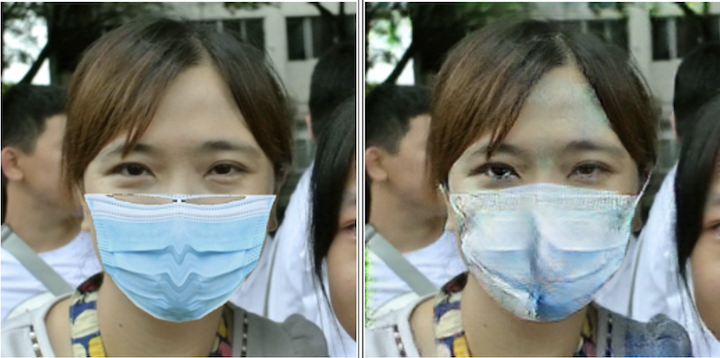, width = 4.4cm}}
   {\epsfig{file = 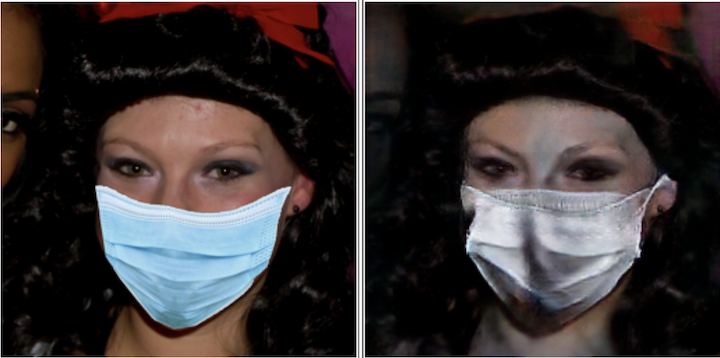, width = 4.4cm}}
   {\epsfig{file = 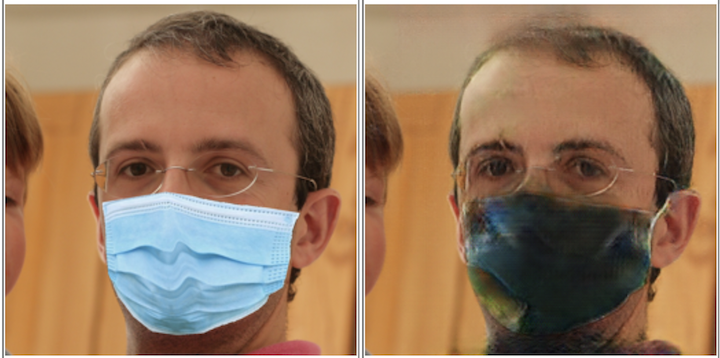, width = 4.4cm}}
   {\epsfig{file = 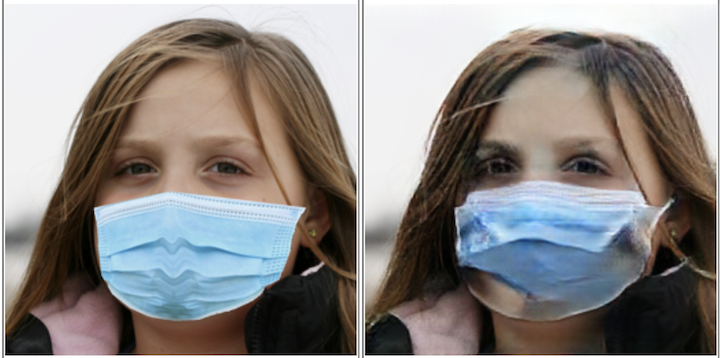, width = 4.4cm}}
   {\epsfig{file = 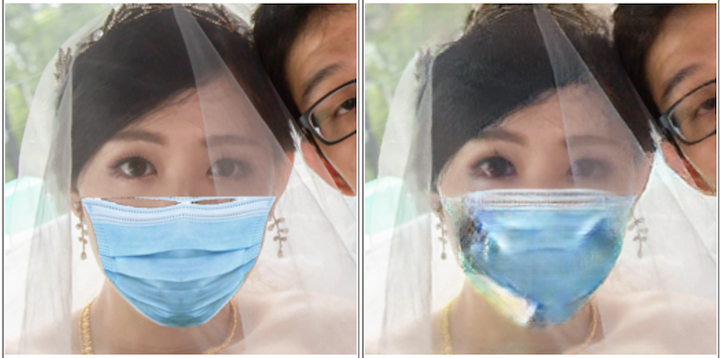, width = 4.4cm}}
  \caption{Results produced by the model at epoch 476. Input and output images are paired side by side.}
  \label{fig:epoch476}
\end{figure*}

In epoch 476, as shown in Figure \ref{fig:epoch476}, all the diversity and details mentioned here are rendered in even more powerful ways. Images such as the top-left and the bottom-right ones in Figure \ref{fig:epoch476} even learned to partly put other occlusions, such as hand or veil, in front of the mask occlusion.

\begin{figure}
\begin{minipage}[c][.258\textheight][c]{.5\textwidth}
  \centering
  \subfloat[]{\includegraphics[width=3.9cm]{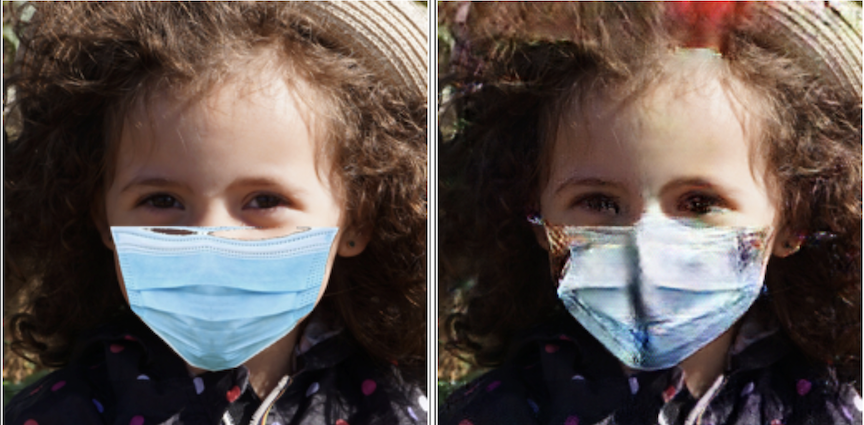}\label{fig:476_noise_hat}}
  \vfill
  \subfloat[]{\includegraphics[width=3.9cm]{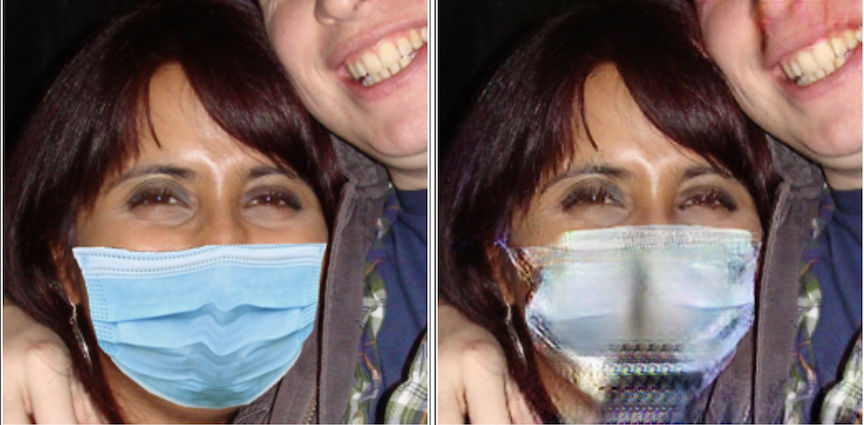}\label{fig:476_noise_scarf}}
  \caption{Noisy output in epoch 476 test result. (a) Red and white blocks in hair. (b) Patterns on the mask.}
  \label{fig:476_noise}
\end{minipage}%
\end{figure}

However, epoch 476 also produced more noise and distortions than epoch 313. We think this is due to overfitting the small training set. Red and white colors in the hair and forehead (Figure \ref{fig:476_noise_hat}) are likely caused by faces wearing not only masks but also Christmas hats appearing in our target training set repeatedly. Patterns on the bottom parts of masks (Figure \ref{fig:476_noise_scarf}) may be caused by a large portion of images with patterned scarves occluding the masks.

\subsection{Comparison with IAMGAN}

\begin{figure}[!ht]
  \vspace{-0.2cm}
  \centering
   {\epsfig{file = 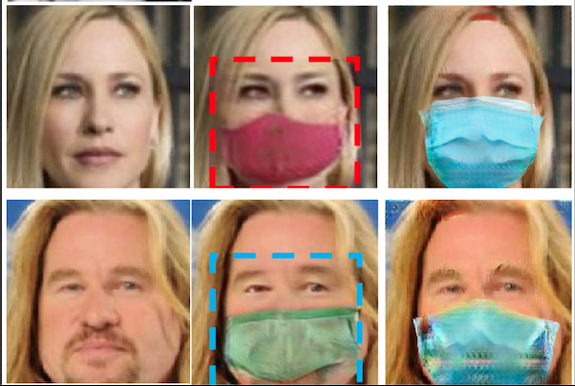, width = 5cm}}
   {\epsfig{file = 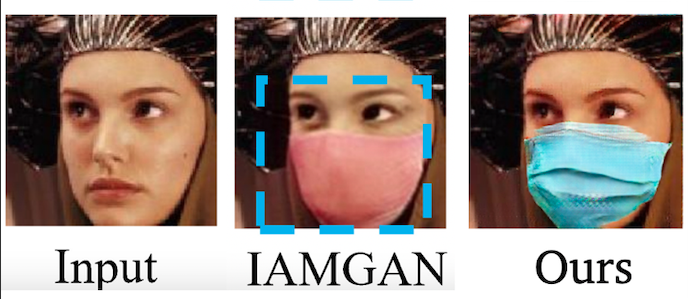, width = 5cm}}
  \caption{Comparing our model with IAMGAN.}
  \label{fig:comp_iamgan}
  \vspace{-0.1cm}
\end{figure}

Both \citet{geng2020masked} and our research used CycleGAN-based methods to turn full faces into masked ones. Our differences include:
\begin{itemize}
    \item IAMGAN uses a multi-layer identity loss, while our NMC loss is pixel-level only. They differ because IAMGAN aims to keep the person's identity after adding a mask, while we want to keep the images exactly the same except the mask region to facilitate both recognition and detection tasks.
    \item IAMGAN always predicts the mask regions, while we utilize ground truth mask regions during training and only predict it during testing.
    \item IAMGAN works on more diverse data, while we have pioneer work on constrained datasets.
    \item IAMGAN turns full faces directly into masked ones, while we require a pre-step and turn fake masks into more realistic ones after the pre-step.
\end{itemize}
Performance scores such as Frechet Inception Distance (FID) \citep{DBLP:journals/corr/HeuselRUNKH17} and Kernel Inception Distance (KID) \citep{binkowski2018demystifying} are usually used to compare the fidelity of different synthesized datasets, but we lack a real-world masked face dataset as the baseline. Datasets highly similar to either IAMGAN's or our training data would unfaithfully push one party's score high. Therefore, we ran our model on some examples shown in the IAMGAN paper \citep{geng2020masked} and demonstrated a qualitative comparison in Figure \ref{fig:comp_iamgan}.

The two models showed similar abilities to retain non-mask regions. Benefiting from the guidance of the superimposed fake masks, our model got more accurate nose bridge positions and occasionally more details such as fabric folds and connecting points between masks and straps. However, IAMGAN offered good fabric and lighting details in many cases, too, and it offered higher diversity in mask colors.

\subsection{Potential Improvements}

We believe a more thorough work on datasets would greatly benefit the results in the future. It would be best to simultaneously achieve the mutual similarity between sets A and B, the size, and the diversity.

Making the two datasets, i.e., A and B, more similar to each other, with masks being the only source of heterogeneity, is an alternative approach to improving attention learning and reducing distortions to non-mask face areas, complementing the extra loss function in Section \ref{sec:penalty}. As mentioned in Section \ref{sec:data}, we have already limited our data to only a subset from MAFA and Unsplash.com. This step is exactly based on the consideration of limiting irrelevant heterogeneity. However, it resulted in a small dataset. While the steps described in Section \ref{subsec:transfer_learning} using larger datasets to get our initial weights before training on the small datasets helped alleviate the dataset size problem, getting a more considerable amount of quality data may provide further improvements.

Future work on datasets may also emphasize increasing mask color/type diversity, but the type diversity should be attempted together with increased model abilities in learning diverse mask shapes. The cycle loss in our model is better at dealing with point-to-point mapping with little shape-changing, so it may be insufficient for the shape diversity. Besides, we only need the A-to-B model, not the opposite direction. Therefore, the single-sided domain mapping proposed by \citet{DBLP:journals/corr/BenaimW17}, with its distance constraints substituting the cycle loss, could be one direction we consider together with an increased mask type diversity. If we retain the two-directional training architecture, the distance constraints may even be used together with the cycle loss.

Besides improving the data, the non-mask penalty loss itself can be improved in two different ways. First, besides calculating the loss based on the improper content change, we may compare the ground truth attention masks directly with the generator-produced attention masks, taking the differences between the two as an extra loss. Second, instead of using a binary tensor indicating whether each pixel is supposed to be changed or not, we may set a finer-weighted penalty that punishes pixel changes farther away from the mask more than those closer to the mask. Such a weighted penalty would allow more room for the model to create realistic details in the transition regions, for example, mask straps and fabric folds. These improvements to the non-mask penalty will further increase the learning stability and reduce improper changes outside the mask.

\section{\uppercase{Conclusions}}

We aimed at turning full face detection/recognition datasets into masked face datasets, supplementing the limited training data for masked face tasks. For this purpose, we proposed a two-step data augmentation method, utilizing \citet{cabani2021maskedface}'s algorithm to warp mask images onto faces as a pre-step to an AttentionGAN-like model that generates more realistically masked faces. We applied multiple improvements to the GAN model training and verified their effectiveness through experimental results. Analyses of our final results showed that the two-step method provided noticeable improvements compared to using a rule-based method alone. Even with the latest advances of the rule-based method by \citet{DBLP:journals/corr/abs-2109-05804}, we still expect an extra I2I step to render the rule-based results with more details, such as irregular region boundaries caused by fabric folds and straps. Our results are also comparable with state-of-the-art NN-only mask generation methods such as IAMGAN, with complementary details. For example, we produced lighting changes and mask stripes or their connecting points missing in IAMGAN results.

While our current model and the generated images can be used in masked face detection or recognition tasks, we have limitations, including patterned noise caused by overfitting small datasets, the remaining face distortions, and the lacking of diversity in mask color and type. Based on discussions about these limitations, we pointed out several directions to generate even better supplemental training data in the future.

\bibliographystyle{apalike}
\bibliography{main}

\end{document}